\documentclass[]{spie}

\usepackage{amsmath,amssymb}
\usepackage{graphicx}
\usepackage{booktabs}
\usepackage{multirow}
\usepackage{microtype}
\usepackage{hyperref}
\usepackage{cite}

\title{SMKC: Sketch Based Kernel Correlation Images for Variable Cardinality Time Series Anomaly Detection}

\author[a]{Haokun Zhou}
\affil[a]{Imperial College London}
\authorinfo{Correspondence: Haokun Zhou, E-mail: haokun.zhou25@imperial.ac.uk}

\begin{document}
\maketitle

\begin{abstract}
Conventional anomaly detection in multivariate time series relies on the assumption that the set of observed variables remains static. In operational environments, however, monitoring systems frequently experience sensor churn. Signals may appear, disappear, or be renamed, creating data windows where the cardinality varies and may include values unseen during training. To address this challenge, we propose SMKC, a framework that decouples the dynamic input structure from the anomaly detector. We first employ permutation-invariant feature hashing to sketch raw inputs into a fixed size state sequence. We then construct a hybrid kernel image to capture global temporal structure through pairwise comparisons of the sequence and its derivatives. The model learns normal patterns using masked reconstruction and a teacher-student prediction objective. Our evaluation reveals that robust log-distance channels provide the primary discriminative signal, whereas cosine representations often fail to capture sufficient contrast. Notably, we find that a detector using random projections and nearest neighbors on the SMKC representation performs competitively with fully trained baselines without requiring gradient updates. This highlights the effectiveness of the representation itself and offers a practical cold-start solution for resource-constrained deployments.
\end{abstract}


\section{Introduction}
Anomaly detection in multivariate time series is a core capability for monitoring and control systems, where rare deviations may indicate faults, attacks, or performance regressions.
Although recent learning-based detectors have advanced performance on fixed-channel benchmarks, they often inherit a strong assumption that the dimensionality and semantics of the input remain stable.
In operational monitoring stacks this assumption is routinely violated.
Instrumentation changes introduce new metrics, retire old sensors, and alter naming conventions; missingness patterns shift with deployment and load; and windows may contain different numbers of variables depending on what is available at the time.
These dynamics create a practically important regime in which each window contains a variable number of channels $C$, variable order is not meaningful, and test-time cardinalities may be unseen during training.

Many widely used multivariate time-series anomaly detectors are not designed for this regime.
Reconstruction-based and attention-based models typically bind parameters to fixed channel indices or learn dependencies over a stable set of variables, which can require padding, retraining, or manual engineering when sensors churn.
Representative examples include USAD, TranAD, and Anomaly Transformer, as well as graph-based detectors that model inter-sensor relations on a fixed node set.\cite{blazquez2021survey,audibert2020usad,tuli2022tranad,xu2022anomalytransformer,deng2021gdn}
In contrast, our goal is to make variable-cardinality handling a property of the representation rather than an architectural constraint of each detector.

This paper proposes SMKC, a two-stage approach to window-level anomaly detection under dynamic $C$ and missingness.
In the first stage, each raw window $(X,M)$ is mapped into a fixed-width hashed state sequence $g_{1:L}$ using signed feature hashing of variable identifiers.\cite{weinberger2009featurehash}
We explicitly encode missingness using a presence stream alongside values, and normalize by the per-time observed count to reduce the trivial dependence of feature magnitude on $C$.
The construction is permutation invariant over variables and does not require a closed vocabulary of variable identities, enabling generalization to previously unseen variables by design.
In the second stage, we construct a kernel image that captures global temporal structure through pairwise comparisons computed over $g$, $\Delta g$, and $|\Delta g|$.
A kernel-aware Transformer encoder learns normal structure using masked kernel reconstruction inspired by masked autoencoding, together with a complementary teacher--student masked prediction objective motivated by momentum distillation.\cite{vaswani2017attention,he2022mae,grill2020byol,caron2021dino}

A central practical finding is that the representation itself can support strong detection even without training a deep model.
Flattening the SMKC kernel image, applying a Johnson--Lindenstrauss random projection, and scoring by kNN yields a training-free detector that is competitive with trained baselines.\cite{johnson1984jl,achlioptas2003rp,knorr1998distanceoutliers}
This observation suggests a deployment spectrum: when training data or compute is limited, a training-free score over a well-designed representation can provide an effective cold-start detector; when resources permit, masked representation learning and validation-time fusion deliver additional gains.

\section{Related Work}
Time-series anomaly detection has been studied extensively, spanning classical statistical and distance-based methods, density estimation, and deep learning approaches.\cite{blazquez2021survey}
Deep models commonly learn normal structure via reconstruction or forecasting and score deviations by error.
USAD stabilizes training with a two-decoder scheme, and recurrent probabilistic models such as OmniAnomaly model temporal uncertainty.\cite{audibert2020usad,su2019omnianomaly}
Transformer-based detectors have become prominent because attention can capture long-range dependencies; TranAD introduces a two-phase reconstruction mechanism and Anomaly Transformer proposes association discrepancy for anomaly scoring.\cite{tuli2022tranad,xu2022anomalytransformer}
While effective in fixed-channel benchmarks, these methods typically assume a stable input dimensionality and consistent channel identity.

Graph-based methods explicitly model inter-variable dependencies and can improve interpretability when sensor identities are stable.
GDN and MTAD-GAT learn graph-structured representations for multivariate anomaly detection.\cite{deng2021gdn,zhao2020mtadgat}
Such approaches, however, presuppose a fixed node set and persistent sensor semantics, which is misaligned with frequent schema changes and unseen variables.

Missingness is pervasive in deployed telemetry.
GRU-D integrates masks and time gaps to handle missing values explicitly.\cite{che2018grud}
However, missing-value handling does not resolve the distinct problem of missing variables and changing cardinality: many architectures still require a fixed-dimensional channel layout.
Our approach treats missingness as a first-class signal through a presence hash stream while removing dependence on variable ordering and identity through permutation-invariant aggregation.

Permutation invariance has been formalized in set learning.
Deep Sets and Set Transformer characterize and implement permutation-invariant functions over unordered collections.\cite{zaheer2017deepsets,lee2019settransformer}
Directly applying set attention to per-variable sequences can be expensive and can reintroduce identity leakage when learned embeddings are tied to variable IDs.
In our experiments, we include DeepSets- and SetTransformer-based detectors as direct permutation-invariant baselines for variable-cardinality anomaly detection.

We instead adopt deterministic sketching via signed feature hashing,\cite{weinberger2009featurehash} which yields a fixed-width representation and naturally supports unseen identities.

A complementary thread represents time series through matrix- or image-like transforms.
Recurrence plots visualize recurrences via pairwise comparisons,\cite{eckmann1987recurrence} and several works encode time series into images for learning.\cite{wang2015timeseriesimages}
In anomaly detection, MSCRED constructs multiscale ``signature matrices'' that capture correlations across sensors.\cite{zhang2019mscred}
SMKC is related in spirit, but differs in that the kernel image is built from a variable-cardinality, permutation-invariant hashed sequence, and it stacks a robust log-distance transform with bounded cosine similarities over temporal derivatives to expose both magnitude-based and directional structure.
Our ablations indicate that the robust log-distance channels provide the dominant signal in the synthetic benchmark, while cosine similarities alone are insufficient. We also include an MSCRED-style signature-matrix baseline to contrast with identity-agnostic representations under sensor churn.

Finally, our scoring and baselines connect to nonparametric anomaly detection.
Distance- and density-based outlier scoring via nearest neighbors is a classical approach,\cite{knorr1998distanceoutliers} and random projection provides a principled compression technique with approximate distance preservation.\cite{johnson1984jl,achlioptas2003rp}
The competitiveness of RandProj-kNN in our results emphasizes that representation design can be a primary lever, with learning acting as an optional refinement.

\section{Methodology}
\subsection{Overview and design goals}
This work targets window-level anomaly detection in multivariate time series when the number of observed variables is not fixed. Each example is a length-$L$ window with $C$ variables and missingness, where $C$ may vary across windows and may take values that are not observed during training. This setting creates two practical requirements that standard detectors often do not satisfy. First, the model must accept variable-cardinality inputs without retraining or resizing layers. Second, the model must avoid relying on variable identities or column order, because real systems frequently add, remove, or rename signals.

We meet these requirements by separating the pipeline into two stages. The first stage converts raw variable-cardinality inputs into a fixed-shape representation using a hashing-based aggregation that is permutation invariant over variables. The second stage learns a representation and anomaly score from that fixed-shape input using masked modeling objectives that encourage contextual reasoning rather than memorization. This separation is central to our approach because it makes dynamic $C$ a property of the representation, not a special case handled inside each baseline architecture.

\subsection{SMKC representation for variable-cardinality inputs}
\subsubsection{From $(X,M)$ to a fixed-width hashed sequence}
Each window provides values $X\in \mathbb{R}^{L\times C}$ and a missingness indicator $M\in\{0,1\}^{L\times C}$. Directly feeding $X$ into a neural network requires a fixed $C$, an ordering of variables, or both. Instead, we map the window into a fixed-width ``hashed state sequence'' $g_{1:L}$, where each $g_t\in \mathbb{R}^{2m}$ and $m$ is a fixed number of hash buckets.

At each time step $t$, we aggregate variable values into buckets using signed feature hashing.
Let $h_v(\cdot)\in\{1,\dots,m\}$ and $s_v(\cdot)\in\{-1,+1\}$ denote the bucket index and sign for the value stream, and $h_p(\cdot)$ and $s_p(\cdot)$ the corresponding functions for the presence stream.
The value sketch $\phi^{(v)}_t\in\mathbb{R}^m$ and presence sketch $\phi^{(p)}_t\in\mathbb{R}^m$ are formed as
\begin{equation}
\phi^{(v)}_{t,k}=\sum_{j=1}^{C} s_v(\mathrm{id}_j)\,X_{t,j}\,\mathbb{I}[h_v(\mathrm{id}_j)=k],\quad
\phi^{(p)}_{t,k}=\sum_{j=1}^{C} s_p(\mathrm{id}_j)\,M_{t,j}\,\mathbb{I}[h_p(\mathrm{id}_j)=k],
\end{equation}
where $\mathrm{id}_j$ is the identifier of variable $j$ and $k$ indexes hash buckets.
We then concatenate and normalize:
\begin{equation}
g_t \;=\;\frac{1}{\sqrt{n_t}}\left[\;\phi^{(v)}_t,\;\lambda(n_t)\,\phi^{(p)}_t\;\right]\in\mathbb{R}^{2m}, \qquad
n_t=\sum_{j=1}^C M_{t,j},
\end{equation}
where $\lambda(n_t)=\min\{a_{\mathrm{pres}}\,n_t,\,1\}$ is a saturating scaling that makes presence informative at small counts without dominating at large counts.

\paragraph{Implementation details: $\lambda(n_t)$ and deterministic hashing.}
In all experiments we set the padding bound to $C_{\max}=16$ (used only by the padding baseline) and define the saturation count
$n_{\mathrm{sat}}=\lceil 0.3\,C_{\max}\rceil=5$.
Accordingly, we set $a_{\mathrm{pres}}=1/n_{\mathrm{sat}}=0.2$ so that $\lambda(n_t)=\min\{0.2\,n_t,\,1\}$, i.e., the presence stream saturates once roughly five variables are observed at a time step.
This choice matches the implementation (\texttt{A\_PRES = 1/ceil(0.3*Cmax)}).

All hash bucket indices and signs are computed deterministically using MD5, so there is no stochastic ``hashing seed''.
Concretely, for a variable identifier $\mathrm{id}$ we define
$h_v(\mathrm{id}) = 1 + (\mathrm{MD5}(\mathrm{id}\Vert \texttt{\#val}) \bmod m)$ and
$s_v(\mathrm{id}) = (-1)^{\mathrm{MD5}(\mathrm{id}\Vert \texttt{\#val\_sign}) \bmod 2}$ for the value stream,
and analogously $h_p(\cdot)$ and $s_p(\cdot)$ using suffixes \texttt{\#pres} and \texttt{\#pres\_sign} for the presence stream.
We also assign variable groups deterministically using $\mathrm{MD5}(\mathrm{id}\Vert \texttt{\#group}) \bmod 4$.
Using independent suffixes decouples the value and presence sketches and ensures that results are reproducible across runs and machines.

This construction is permutation invariant over variables, produces a constant input size regardless of $C$, and uses only deterministic hashes of variable IDs. As a result, the downstream model supports dynamic $C$ natively and does not require an explicit variable ordering.

The hash width $m$ trades off information retention against computation. Increasing $m$ reduces collisions in the signed feature-hash sketch, but it also increases the dimensionality of $g_t\in\mathbb{R}^{2m}$ and thus the cost of downstream kernel computations, which scale linearly with $m$ for fixed window length $L$. Because collisions are central to sketch fidelity, we explicitly evaluate sensitivity to $m$ in Section~\ref{sec:msweep}. In the main experiments we use $m=128$ as a conservative default that yields low collision rates while keeping the representation compact.

\subsubsection{From hashed sequence to a hybrid kernel image}
A fixed-width sequence is sufficient for many sequence models, but our approach uses an additional kernel-matrix view that exposes global temporal structure.
Given the hashed sequence $g_{1:L}$, we compute three derived sequences: the original sequence $g$, its first difference $\Delta g$ (with $\Delta g_1=0$), and the elementwise absolute difference $|\Delta g|$.
For each sequence $z\in\{g,\Delta g,|\Delta g|\}$ we compute (i) a cosine similarity matrix and (ii) a robust log-distance matrix.

For cosine similarity, we $\ell_2$-normalize each time step and compute
\begin{equation}
\mathrm{Cos}(z)_{i,j} \;=\; \tfrac{1}{2}\left(1+\frac{z_i^\top z_j}{\|z_i\|_2\,\|z_j\|_2}\right)\in[0,1].
\end{equation}
For the robust log-distance channel, we compute squared distances $D_{i,j}=\|z_i-z_j\|_2^2$ and a robust bandwidth
\begin{equation}
\sigma(z) \;=\; \mathrm{median}_{i<j}\;\|z_i-z_j\|_2,
\end{equation}
then form
\begin{equation}
\mathrm{LogDist}(z)_{i,j} \;=\; \log\!\left(1+\frac{D_{i,j}}{2\sigma(z)^2}\right).
\end{equation}
Following the implementation, we also compute a scalar scale token $\tau=\tanh(\log(\sigma(g)))$, which conditions the encoder on the within-window distance scale.

Stacking the six matrices
$\{\mathrm{Cos}(g),\mathrm{Cos}(\Delta g),\mathrm{Cos}(|\Delta g|),\mathrm{LogDist}(g),\mathrm{LogDist}(\Delta g),\mathrm{LogDist}(|\Delta g|)\}$
yields a fixed tensor of shape $6\times L\times L$ that we refer to as the SMKC hybrid image.
The cosine channels are bounded and emphasize directional similarity and smoothness, while the log-distance channels emphasize magnitude-based deviations with robust compression of large distances.
Although we include both families for generality, our ablation study in Section~\ref{sec:ablations} shows that, in the synthetic benchmark, the robust log-distance family is the dominant contributor: cosine-only representations collapse, whereas a log-distance-only variant is both competitive and slightly stronger, offering a simplified three-channel alternative.

\begin{figure*}[t]
\centering
\includegraphics[width=\textwidth]{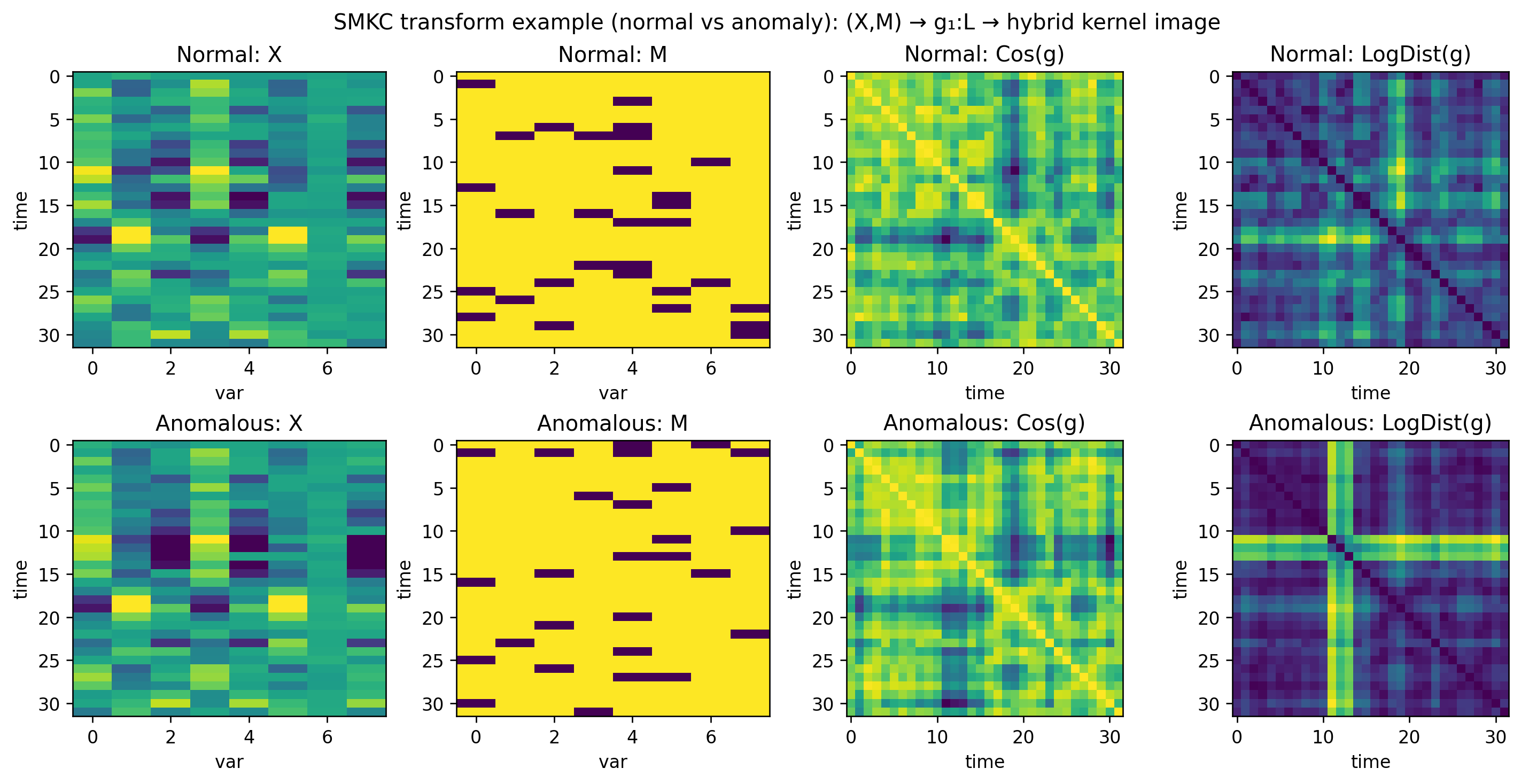}
\caption{SMKC transform example (normal vs. anomalous). Starting from a window with values $X$ and missingness mask $M$, we construct the fixed-width hashed state sequence $g_{1:L}$ and then compute kernel-view channels. Shown are two representative channels derived from $g$: cosine similarity $\mathrm{Cos}(g)$ and robust log-distance $\mathrm{LogDist}(g)$. While missingness can obscure localized changes in $X$, the kernel view reveals structured deviations that are more amenable to masked modeling and window-level scoring.}
\label{fig:smkc-transform}
\end{figure*}

\begin{figure*}[t]
\centering
\includegraphics[width=\textwidth]{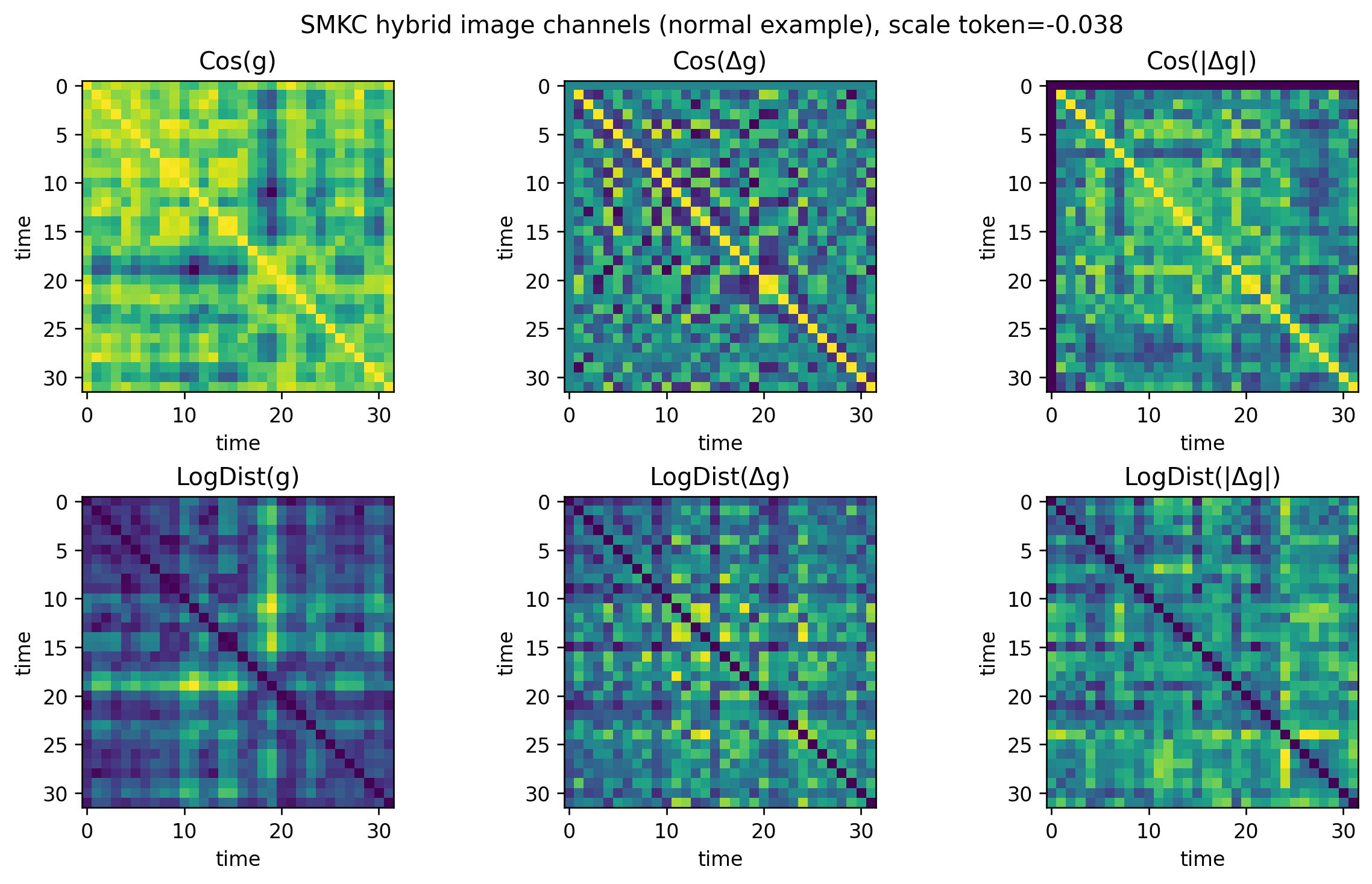}
\caption{Six-channel SMKC hybrid kernel image for a normal window. We stack cosine similarity and robust log-distance matrices computed on the hashed sequence $g$, its first difference $\Delta g$, and the absolute difference $|\Delta g|$. The cosine channels emphasize bounded directional similarity, whereas the log-distance channels emphasize magnitude-based deviations with robust compression. The scale token summarizes the robust bandwidth used in the log-distance transform and conditions the encoder on the window’s intrinsic distance scale.}
\label{fig:smkc-hybrid-channels}
\end{figure*}

Figure~\ref{fig:smkc-transform} visualizes the end-to-end SMKC transform on representative normal and anomalous windows.
Although the raw values $X$ can appear only subtly perturbed under missingness, the kernel-view channels computed from the hashed sequence $g_{1:L}$ often exhibit coherent structural deviations.
Figure~\ref{fig:smkc-hybrid-channels} further decomposes the hybrid kernel image into its six channels, illustrating how cosine similarity and robust log-distance provide complementary views of static and dynamic temporal structure.

\subsection{Addressing the $O(L^2 d)$ bottleneck: scalable HYB representations}

Our baseline detector is the SMKC RandProj--kNN density estimator: we flatten a window representation,
project it to 256 dimensions with a fixed random matrix, and compute anomaly scores from the average
cosine distance to the $K{=}20$ nearest neighbors in a normal-only reference set.
To isolate representation cost, we keep the detector fixed and ablate only the representation.

\paragraph{Baseline HYB image ($O(L^2 d)$).}
Given a hashed state sequence $g_{1:L}\in\mathbb{R}^{L\times d}$ (with $d{=}2m$),
the classic HYB representation materializes an $L\times L$ pairwise “hybrid image”
with $K{=}6$ channels:
three cosine-similarity channels (raw, $\Delta$, $|\Delta|$) mapped to $[0,1]$,
and three log-distance channels (raw, $\Delta$, $|\Delta|$) using per-window median distance
normalization.
This has compute complexity $O(K L^2 d)$ and memory $O(K L^2)$, and its flattened feature
dimension is $K L^2$.

\paragraph{Constant-factor ablations (still $O(L^2 d)$).}
We test standard constant-factor wins:
(i) channel pruning \texttt{log3} (keep only the three log-distance channels) and \texttt{base2}
(keep only raw cosine + raw log-distance),
(ii) projecting the sequence to $d'\in\{128,64\}$ \emph{before} pairwise computations (reduces the
pairwise inner-product cost but does not change the final feature dimension),
and (iii) temporal patching (mean-pool time to $L'=16$ and then build the full HYB image).

\paragraph{Linear-in-$L$ scaling fixes: \texttt{bandfeat} and \texttt{anchorfeat}.}
To remove the $L\times L$ materialization entirely, we introduce two structured feature maps:

\begin{itemize}
\item \textbf{Band-diagonal features (\texttt{bandfeat}).}
Instead of all pairs, we compute only the $w$ near-diagonal lags (diagonals) for each channel.
This yields a tensor of shape $(K, w, L)$ and reduces compute to $O(K\,L\,w\,d)$ and memory to
$O(K\,L\,w)$.

\item \textbf{Time-to-anchor features (\texttt{anchorfeat}).}
We select $r$ anchor timesteps and compute similarity/distance from each time index to each anchor,
giving a tensor $(K, L, r)$ with compute $O(K\,L\,r\,d)$ and memory $O(K\,L\,r)$.
\end{itemize}

Both representations preserve pairwise structure while scaling linearly in $L$ when $w,r\ll L$.

\paragraph{Protocol.}
We follow the synthetic holdout-$C$ setup used throughout this study:
train on normal-only windows for $C\in\{1,2,4,8\}$ and test on held-out $C\in\{3,6,12,16\}$ with
an anomaly rate of 0.10. We use $L{=}64$, $m{=}128$ (thus $d{=}256$), and report AUROC/AUPRC along
with wall-clock runtime (representation build + RandProj--kNN scoring), peak allocated GPU memory,
and a normalized complexity proxy relative to \texttt{HYB-full(img)} (Table~\ref{tab:smkc_scaling_ablation}).

\paragraph{Results and takeaways.}
The best overall variant is \texttt{HYB-log3+bandfeat(w=8)}, which \emph{improves} AUPRC over the
full $6$-channel HYB image while reducing the analytical compute proxy by $16\times$.
Specifically, \texttt{bandfeat} attains AUPRC 0.578 (vs.\ 0.554 for \texttt{HYB-full(img)})
at 0.221s total time (vs.\ 0.763s), driven primarily by a much smaller flattened feature dimension
(1,536 vs.\ 24,576), which makes the RandProj--kNN scoring stage $\approx 10\times$ faster.
A more aggressive band width \texttt{w=4} further reduces the compute proxy to $32\times$ and
improves speed (0.180s total) while maintaining AUPRC 0.565.
Among quadratic ($L\times L$) representations, \texttt{HYB-log3(img)} is the strongest constant-factor
baseline (AUPRC 0.565) but still scales as $O(L^2 d)$.
Temporal patching to $L'=16$ delivers a $16\times$ compute proxy reduction, but substantially degrades
AUPRC (0.464), indicating that aggressive time pooling discards discriminative structure.
Finally, the raw-sequence lower bound \texttt{SEQ(gseq)} performs poorly (AUPRC 0.196), confirming
that explicit pairwise structure is essential for this detector family.

\begin{table}[t]
\centering
\small
\begin{tabular}{l l r r r r r r}
\toprule
Variant & Rep. & AUPRC $\uparrow$ & AUROC $\uparrow$ & Time (s) $\downarrow$ & Peak MB $\downarrow$ & Rel.\ cost $\downarrow$ & Feat.\ dim \\
\midrule
HYB-log3+bandfeat(w=8) & bandfeat & \textbf{0.578} & 0.738 & 0.221 & 101.0 & 0.0625 & 1536 \\
HYB-log3+bandfeat(w=4) & bandfeat & 0.565 & 0.729 & 0.180 & 100.6 & 0.0312 & 768 \\
HYB-log3(img) & img & 0.565 & 0.716 & 0.300 & 106.5 & 0.5000 & 12288 \\
HYB-full+proj128(img) & img & 0.555 & 0.719 & 0.544 & 112.3 & 0.5000 & 24576 \\
\textit{Baseline: HYB-full(img)} & img & 0.554 & 0.715 & 0.763 & 114.2 & 1.0000 & 24576 \\
HYB-full+proj64(img) & img & 0.551 & 0.728 & 0.440 & 109.2 & 0.2500 & 24576 \\
HYB-log3+anchor(r=16) & anchorfeat & 0.545 & 0.724 & \textbf{0.151} & 102.9 & 0.1250 & 3072 \\
HYB-log3+anchor(r=8) & anchorfeat & 0.528 & 0.720 & 0.172 & 101.5 & 0.0625 & 1536 \\
HYB-base2(img) & img & 0.484 & 0.698 & 0.274 & 105.2 & 0.3333 & 8192 \\
HYB-full+down16(img) & img & 0.464 & 0.697 & 0.167 & 31.8 & 0.0625 & 1536 \\
SEQ(gseq) & seq & 0.196 & 0.625 & 0.480 & -- & 0.0026 & 16384 \\
\bottomrule
\end{tabular}
\caption{SMKC scaling ablation on the synthetic holdout-$C$ protocol ($L=64$, $m=128$ so $d=256$).
Rel.\ cost is the analytical complexity proxy normalized by \textit{HYB-full(img)} (lower is better).
Time includes representation construction + RandProj--kNN scoring.}
\label{tab:smkc_scaling_ablation}
\end{table}

\paragraph{End-to-end scaling is dominated by the encoder at large $L$.}
While Table~\ref{tab:smkc_scaling_ablation} isolates training-free detectors (representation + RandProj--kNN),
the end-to-end cost of the full pipeline (representation + encoder) is governed primarily by the encoder
as $L$ increases.
Table~\ref{tab:e2e_time} summarizes per-window wall-clock time at $m{=}128$ (patch size $P{=}2$).
At $L{=}32$, the complete pipeline runs in roughly $0.10$--$0.13$ ms/window across both backbones.
As $L$ grows, attention scaling dominates: at $L{=}128$, standard self-attention rises to $6.328$ ms/window,
whereas linear attention remains $1.203$ ms/window (a $\approx 5.26\times$ reduction).
Training-time follows the same pattern: at $L{=}128$, the train step is $4.955$ ms/window with MHSA
and $3.031$ ms/window with linear attention.

\begin{table}[t]
\centering
\small
\begin{tabular}{l r r r r}
\toprule
Encoder backbone & Inference ($L{=}32$) & Inference ($L{=}128$) & Train step ($L{=}32$) & Train step ($L{=}128$) \\
\midrule
Self-attention (MHSA) & 0.102 & 6.328 & 0.268 & 4.955 \\
Linear attention      & 0.125 & 1.203 & 0.323 & 3.031 \\
\bottomrule
\end{tabular}
\caption{End-to-end wall-clock time for representation + encoder at $m{=}128$ (patch size $P{=}2$),
reported in milliseconds per window.}
\label{tab:e2e_time}
\end{table}

\subsection{Model architecture and scoring}
\subsubsection{Kernel-aware encoder}
The SMKC hybrid image is processed by a convolutional stem followed by a patch embedding layer and a Transformer encoder. The patch embedding converts the $L\times L$ image into a grid of patch tokens. We then add a positional encoding that reflects the semantics of a kernel matrix rather than a generic image. Specifically, each token receives components corresponding to its row index, column index, and lag index. The lag index depends on the absolute difference between row and column positions, which aligns with the intuition that near-diagonal regions represent short temporal lags while far-from-diagonal regions represent long lags.

In addition to patch tokens, the encoder receives a learned classification token that serves as the window embedding and a scale token that embeds the SMKC scale scalar. The scale token acts as a context variable that allows the network to modulate attention and reconstruction behavior based on the window's intrinsic distance scale.

\subsubsection{Masked kernel reconstruction objective}
To learn normal structure without anomaly labels, we train a masked modeling objective on the SMKC image. During training, we mask a large fraction of patches using structured mask patterns that preserve the symmetry of kernel matrices. Mask patterns are sampled from a mixture that includes random symmetric masks as well as masks that remove contiguous rows and columns, banded regions relative to the diagonal, and block regions. This mixture prevents the model from overfitting to a single masking pattern and encourages it to capture both local and long-range dependencies.
A component ablation in Section~\ref{sec:ablations} shows that restricting to random masks alone reduces AUPRC, supporting the use of kernel-aware structured masking.

A cross-attention decoder reconstructs the masked patches from the visible encoder tokens. Because the hybrid image channels have known ranges and constraints, we apply channel-appropriate output activations and include light structural regularizers, such as penalties for breaking symmetry and penalties that keep the diagonal consistent with the expected self-similarity and self-distance values. These constraints reduce degenerate reconstructions that could achieve low loss while violating kernel structure.

\subsubsection{Masked prediction objective and ensemble fusion}
In addition to reconstruction, we train a masked prediction objective inspired by teacher--student self-distillation. A teacher encoder observes the full input while a student encoder observes only the visible patches. A predictor network then maps the student representation to the teacher's representations at masked locations. The teacher is updated by exponential moving average, which stabilizes training and encourages invariance.

The teacher and student share the same kernel-aware encoder architecture and are initialized identically.
The teacher parameters are updated by exponential moving average (EMA) of the student parameters with a momentum that increases linearly from $0.996$ to $0.9999$ over training steps.
The predictor consists of a 4-layer cross-attention decoder and is trained to match the teacher’s patch-token representations at masked locations using a SmoothL1 loss, with an auxiliary SmoothL1 loss on the CLS token weighted by $0.25$.
Mask patterns are sampled from the same symmetric kernel-aware mixture used for reconstruction.

Rather than reporting masked prediction as a separate detector, we use it as a complementary signal to reconstruction. Reconstruction emphasizes pixel-level fidelity of the kernel image, while masked prediction emphasizes consistency of learned representations. These objectives capture different aspects of normality, and their combination improves robustness across regimes and across $C$.

\subsubsection{Anomaly score}
At inference time, we compute two types of signals. The first signal is a density-to-normal score computed by comparing the normalized embedding of a test window to embeddings of training normal windows using k-nearest neighbors. This score detects windows that lie far from the normal manifold. The second signal is a surprise score computed from masked reconstruction or masked prediction errors, averaged across multiple random masks to reduce variance. Both signals are robustly standardized, then combined by a fixed weighted sum.

For the ensemble variant, we learn a lightweight fusion model on a validation set. The fusion model takes as input the density and surprise components from both reconstruction and masked prediction and outputs a calibrated anomaly score. Importantly, this fusion step does not change the representation network. It plays the same role as validation-based calibration commonly used in anomaly detection to choose operating points and combine complementary detectors.

\subsection{Synthetic data generation and interpretation of settings}
The synthetic benchmark is designed to stress variable-cardinality behavior while maintaining interpretability. Each window is generated from latent dynamics and variable-dependent mixing coefficients, then perturbed by controlled anomalies and missingness.

Each variable is assigned a stable identifier and mapped deterministically into one of four groups. Group membership determines how the variable depends on two latent factors: two groups correspond to positive and negative coupling to the first factor, one group corresponds to coupling to the second factor, and one group is noise-only. Training and evaluation use disjoint pools of variable identifiers, preventing models from memorizing identity-specific behavior. Latent factors evolve under regime-specific autoregressive parameters with matched stationary variance. Missingness is applied independently per entry with a fixed probability, while ensuring at least one observed variable per time step.

Anomalies are injected by sampling one anomaly type and applying it over a short segment. The suite includes latent-level perturbations (e.g., factor spikes, coupling changes), observation-level perturbations (e.g., sparse spikes, channel reassignment), lag-copy segments, and regime switching within a window. To avoid label noise from ``invisible'' anomalies that cannot manifest under certain sampled variable compositions, anomaly windows are constrained to include required variable groups, ensuring that anomalous labels correspond to observable changes in $X$.

\subsection{Baselines and dynamic input handling}
Baselines span classical anomaly detectors, sequence reconstruction and attention-based neural models on fixed-dimensional hashed sequences, permutation-invariant set encoders for variable-cardinality inputs, and image-style correlation baselines.

Classical detectors are trained on pooled window-level statistics and include kNN distance, Isolation Forest, LOF, and one-class SVM \cite{knorr1998distanceoutliers,liu2008isolationforest,breunig2000lof,scholkopf2001ocsvm}.
Neural methods that require a fixed feature dimension are adapted to operate on the hashed sequence $g_{1:L}$, thereby isolating architectural effects from variable-cardinality handling; these include GRU autoencoders, masked autoencoding, TranAD, Anomaly Transformer, and USAD \cite{audibert2020usad,tuli2022tranad,xu2022anomalytransformer}.
We additionally evaluate two set-based baselines that operate directly on variable-cardinality inputs: DeepSets-GRU(Set) and SetTransformer-GRU(Set). Both encode each variable trajectory with a GRU and aggregate variable embeddings using permutation-invariant operators \cite{zaheer2017deepsets,lee2019settransformer}.
As a common practical workaround, we also include a padding-based GRU autoencoder that pads values and masks to a maximum $C_{\max}$.
We further include an MSCRED(Signature) baseline, which constructs correlation-style signature matrices and applies a convolutional detector, representing image-based multivariate correlation modeling \cite{zhang2019mscred}.
Finally, we evaluate a training-free baseline that flattens the SMKC image, applies a random projection, and scores anomalies via kNN (RandProj-kNN) \cite{johnson1984jl,achlioptas2003rp,knorr1998distanceoutliers}.

\section{Experiments}
We evaluate two protocols that stress generalization under dynamic cardinality.
In \emph{in\_dist\_C}, training and testing share $C\in\{1,2,3,4,6,8,12,16\}$.
In \emph{holdout\_C}, training uses $C\in\{1,2,4,8\}$ while testing uses unseen $C\in\{3,6,12,16\}$. Unless otherwise stated, we set the hash width to $m=128$; to assess collision sensitivity we additionally sweep $m\in\{32,64,128,256,512\}$ for the training-free RandProj-kNN(SMKC) detector under holdout\_C.

All representation-learning models are trained on normal-only windows.
Validation windows contain anomalies at a controlled rate and are used for hyperparameter selection and, for the ensemble variant, score calibration.
We report mean and standard deviation over three random seeds and evaluate multiple test anomaly rates in $\{0.01,0.05,0.10,0.20\}$, emphasizing AUPRC as the primary metric under imbalance, while also reporting AUROC and operating-point metrics.

\section{Results and Interpretation}
Unless otherwise stated, results in this section are reported at test anomaly rate 0.10 and aggregated over three random seeds.
For the hash-width sweep in Section~\ref{sec:msweep}, we report mean and standard deviation over seven seeds to obtain stable uncertainty estimates.
For the component ablation in Section~\ref{sec:ablations}, we report relative effects under a reduced training budget to enable a broad sweep of variants; we emphasize trends rather than absolute values.

\subsection{Overall detection performance}
Table~\ref{tab:avg-metrics} summarizes performance averaged over the tested $C$ values.
SMKC-ENS(HYB) achieves the strongest ranking performance in both protocols, yielding the highest average AUPRC and AUROC, with the margin being most pronounced under holdout\_C where test cardinalities are unseen during training.
Permutation-invariant set baselines (DeepSets-GRU(Set) and SetTransformer-GRU(Set)) are highly competitive, matching SMKC-Recon(HYB) closely and slightly improving TPR@1\%FPR under in\_dist\_C, which suggests they can be attractive when the operating point emphasizes extremely low false-alarm regimes.
MSCRED(Signature) performs substantially worse in this churn setting, consistent with the fact that correlation-matrix baselines implicitly assume stable channel identities and consistent matrix semantics across windows.

A notable practical outcome is that RandProj-kNN(SMKC), which does not require representation learning, remains competitive with trained neural baselines.
This suggests that the SMKC kernel image exposes substantial discriminative structure even before training, and motivates a cold-start deployment option in which a training-free detector can be used immediately while more expensive learning and calibration can be added as resources permit.

\begin{table}[t]
\centering
\caption{Average performance over tested $C$ values at anomaly rate 0.10. Higher is better.}
\label{tab:avg-metrics}
\begin{tabular}{lcccccc}
\toprule
\multirow{2}{*}{Model} & \multicolumn{3}{c}{in\_dist\_C} & \multicolumn{3}{c}{holdout\_C} \\
\cmidrule(lr){2-4}\cmidrule(lr){5-7}
& AUPRC & AUROC & TPR@1\%FPR & AUPRC & AUROC & TPR@1\%FPR \\
\midrule
SMKC-ENS(HYB) & \textbf{0.549} & \textbf{0.737} & 0.431 & \textbf{0.577} & \textbf{0.755} & \textbf{0.454} \\
SMKC-Recon(HYB) & 0.542 & 0.726 & 0.430 & 0.561 & 0.742 & 0.446 \\
RandProj-kNN(SMKC) & 0.531 & 0.719 & 0.418 & 0.555 & 0.728 & 0.448 \\
Pad+Mask GRU-AE(Cmax) & 0.539 & 0.725 & 0.425 & 0.555 & 0.738 & 0.436 \\
GRU-AE(gseq) & 0.533 & 0.716 & 0.420 & 0.555 & 0.729 & 0.442 \\
DeepSets-GRU(Set) & 0.543 & 0.719 & \textbf{0.435} & 0.559 & 0.727 & 0.450 \\
SetTransformer-GRU(Set) & 0.544 & 0.720 & \textbf{0.437} & 0.559 & 0.727 & 0.453 \\
MSCRED(Signature) & 0.324 & 0.690 & 0.117 & 0.319 & 0.684 & 0.132 \\
AnomalyTransformer(gseq) & 0.522 & 0.710 & 0.411 & 0.534 & 0.725 & 0.407 \\
TS-MAE(gseq) & 0.449 & 0.693 & 0.311 & 0.484 & 0.707 & 0.341 \\
StatsPool-kNN & 0.267 & 0.632 & 0.126 & 0.330 & 0.668 & 0.177 \\
IsolationForest(stats) & 0.305 & 0.636 & 0.167 & 0.307 & 0.654 & 0.154 \\
\bottomrule
\end{tabular}
\end{table}

\subsection{Effect of variable cardinality}
Tables~\ref{tab:auprc-indist} and~\ref{tab:auprc-holdout} report AUPRC$_\text{all}$ as a function of $C$ at anomaly rate 0.10.
Under in\_dist\_C, performance is lowest at very small $C$, where each window provides limited structured evidence and missingness can dominate.
In this regime, SMKC-based methods remain substantially more stable than sequence-only masked autoencoding, suggesting that the kernel view and kernel-aware masking more effectively exploit global temporal structure.
As $C$ increases, most learned methods improve, but SMKC-ENS(HYB) remains best or tied-best across most cardinalities.

Under holdout\_C, all test $C$ values are unseen during training by construction.
Despite this shift, SMKC-ENS(HYB) remains best for $C\in\{6,12,16\}$ and shows its largest margin at $C=16$, consistent with the interpretation that complementary density and surprise signals become particularly useful when more structured variables are present.
The stability of RandProj-kNN across protocols further indicates that the representation captures geometry that transfers across unseen cardinalities even without training.

\begin{table}[t]
\centering
\scriptsize
\caption{AUPRC$_\text{all}$ vs. $C$ under in\_dist\_C at anomaly rate 0.10 (mean $\pm$ std over 3 seeds).}
\label{tab:auprc-indist}
\begin{tabular}{lcccccccc}
\toprule
Model & 1 & 2 & 3 & 4 & 6 & 8 & 12 & 16 \\
\midrule
SMKC-ENS(HYB) & 0.478$\pm$0.033 & 0.542$\pm$0.031 & 0.578$\pm$0.109 & 0.518$\pm$0.117 & 0.590$\pm$0.021 & 0.546$\pm$0.072 & 0.536$\pm$0.047 & 0.605$\pm$0.067 \\
SMKC-Recon(HYB) & 0.493$\pm$0.036 & 0.535$\pm$0.041 & 0.574$\pm$0.105 & 0.526$\pm$0.114 & 0.574$\pm$0.032 & 0.524$\pm$0.058 & 0.513$\pm$0.048 & 0.592$\pm$0.083 \\
RandProj-kNN(SMKC) & 0.475$\pm$0.043 & 0.515$\pm$0.058 & 0.570$\pm$0.087 & 0.525$\pm$0.111 & 0.567$\pm$0.045 & 0.517$\pm$0.066 & 0.506$\pm$0.055 & 0.569$\pm$0.085 \\
Pad+Mask GRU-AE(Cmax) & 0.475$\pm$0.045 & 0.525$\pm$0.021 & 0.571$\pm$0.108 & 0.527$\pm$0.091 & 0.565$\pm$0.044 & 0.544$\pm$0.050 & 0.521$\pm$0.050 & 0.584$\pm$0.074 \\
GRU-AE(gseq) & 0.470$\pm$0.039 & 0.518$\pm$0.025 & 0.570$\pm$0.107 & 0.520$\pm$0.109 & 0.567$\pm$0.046 & 0.541$\pm$0.056 & 0.503$\pm$0.055 & 0.576$\pm$0.064 \\
AnomalyTransformer(gseq) & 0.455$\pm$0.031 & 0.494$\pm$0.025 & 0.571$\pm$0.105 & 0.511$\pm$0.111 & 0.560$\pm$0.052 & 0.530$\pm$0.056 & 0.487$\pm$0.051 & 0.568$\pm$0.049 \\
TS-MAE(gseq) & 0.296$\pm$0.016 & 0.442$\pm$0.034 & 0.514$\pm$0.109 & 0.408$\pm$0.094 & 0.522$\pm$0.055 & 0.474$\pm$0.067 & 0.437$\pm$0.029 & 0.498$\pm$0.068 \\
\bottomrule
\end{tabular}
\end{table}

\begin{table}[t]
\centering
\scriptsize
\caption{AUPRC$_\text{all}$ vs. $C$ under holdout\_C at anomaly rate 0.10 (mean $\pm$ std over 3 seeds).}
\label{tab:auprc-holdout}
\begin{tabular}{lcccc}
\toprule
Model & 3 & 6 & 12 & 16 \\
\midrule
SMKC-ENS(HYB) & 0.574$\pm$0.110 & 0.582$\pm$0.023 & 0.532$\pm$0.052 & 0.621$\pm$0.071 \\
SMKC-Recon(HYB) & 0.579$\pm$0.098 & 0.573$\pm$0.048 & 0.511$\pm$0.042 & 0.580$\pm$0.068 \\
RandProj-kNN(SMKC) & 0.571$\pm$0.089 & 0.570$\pm$0.044 & 0.507$\pm$0.055 & 0.571$\pm$0.087 \\
Pad+Mask GRU-AE(Cmax) & 0.570$\pm$0.110 & 0.565$\pm$0.045 & 0.509$\pm$0.052 & 0.577$\pm$0.073 \\
GRU-AE(gseq) & 0.571$\pm$0.106 & 0.568$\pm$0.047 & 0.503$\pm$0.055 & 0.577$\pm$0.064 \\
AnomalyTransformer(gseq) & 0.554$\pm$0.096 & 0.546$\pm$0.037 & 0.486$\pm$0.051 & 0.550$\pm$0.040 \\
TS-MAE(gseq) & 0.495$\pm$0.118 & 0.519$\pm$0.055 & 0.425$\pm$0.019 & 0.497$\pm$0.072 \\
\bottomrule
\end{tabular}
\end{table}

\subsection{Sensitivity to anomaly prevalence}
Table~\ref{tab:auprc-rate} reports AUPRC averaged over tested $C$ values as a function of the test anomaly rate.
AUPRC generally increases as anomalies become less rare, and the ranking among the strongest methods remains stable.
SMKC-ENS(HYB) is consistently best or tied-best across rates, indicating that the learned fusion does not overfit to a single class-imbalance regime.
RandProj-kNN(SMKC) remains competitive across rates without training, strengthening the implication that the representation design captures transferable structure.

\begin{table}[t]
\centering
\caption{AUPRC$_\text{all}$ averaged over tested $C$ values vs. test anomaly rate.}
\label{tab:auprc-rate}
\begin{tabular}{lcccc}
\toprule
Model & 0.01 & 0.05 & 0.10 & 0.20 \\
\midrule
SMKC-ENS(HYB) & 0.459 & 0.504 & 0.549 & 0.619 \\
SMKC-Recon(HYB) & 0.463 & 0.497 & 0.542 & 0.611 \\
RandProj-kNN(SMKC) & 0.436 & 0.481 & 0.531 & 0.601 \\
Pad+Mask GRU-AE(Cmax) & 0.449 & 0.487 & 0.539 & 0.610 \\
GRU-AE(gseq) & 0.444 & 0.486 & 0.533 & 0.607 \\
AnomalyTransformer(gseq) & 0.431 & 0.475 & 0.522 & 0.600 \\
TS-MAE(gseq) & 0.279 & 0.365 & 0.449 & 0.531 \\
StatsPool-kNN & 0.101 & 0.175 & 0.267 & 0.376 \\
\bottomrule
\end{tabular}
\end{table}

\subsection{Sensitivity to hash width $m$ and collision rates}
\label{sec:msweep}
A reviewer concern is that signed feature hashing can lose information through bucket collisions, and that the choice of hash width $m$ may materially affect detection. To quantify this effect, we sweep $m\in\{32,64,128,256,512\}$ and evaluate the training-free RandProj-kNN(SMKC) detector under the holdout\_C protocol at anomaly rate 0.10. In addition to detection metrics, we report the empirical collision fraction for both the value-hash and the presence-hash streams, computed per window as one minus the ratio of unique bucket indices to the number of variables and then averaged across windows.
Because this sweep is designed to isolate collision effects and to support additional seeds, it is conducted as a targeted study and should be interpreted primarily in terms of trends across $m$.

Table~\ref{tab:msweep} shows that collision rates decrease monotonically as $m$ increases, from roughly $4\%$ at $m=32$ to below $0.3\%$ at $m=512$. Despite this large reduction in collisions, AUPRC and AUROC remain essentially unchanged within uncertainty: the best mean AUPRC occurs at $m=128$, but differences across $m$ are small relative to the standard deviation across seeds. These results indicate that, in this benchmark, performance saturates once collisions are modest and that detection is not bottlenecked by residual collisions in the tested range. Practically, because the cost of computing kernel channels scales linearly with $m$ for fixed $L$, the observed insensitivity supports using moderate hash widths (e.g., $m=64$ or $128$) to reduce memory and compute without sacrificing accuracy.

\begin{table}[t]
\centering
\caption{Sensitivity of RandProj-kNN(SMKC) to hash width $m$ under holdout\_C at anomaly rate 0.10. Collision rates are the mean fraction of variables that share a bucket in each stream (averaged over windows). Metrics are mean $\pm$ std over seven seeds.}
\label{tab:msweep}
\begin{tabular}{ccccc}
\toprule
$m$ & Value collision & Presence collision & AUPRC & AUROC \\
\midrule
32  & 0.0406 & 0.0402 & 0.5799$\pm$0.0327 & 0.7323$\pm$0.0338 \\
64  & 0.0208 & 0.0205 & 0.5813$\pm$0.0356 & 0.7377$\pm$0.0337 \\
128 & 0.0114 & 0.0099 & \textbf{0.5844$\pm$0.0356} & 0.7362$\pm$0.0303 \\
256 & 0.0051 & 0.0054 & 0.5831$\pm$0.0373 & 0.7360$\pm$0.0301 \\
512 & 0.0026 & 0.0023 & 0.5836$\pm$0.0372 & 0.7357$\pm$0.0308 \\
\bottomrule
\end{tabular}
\end{table}

\subsection{Component ablation of the SMKC representation and training design}
\label{sec:ablations}
To attribute performance to specific design choices, we ablate components of SMKC-Recon(HYB) under the holdout\_C protocol at anomaly rate 0.10.
To enable a broad sweep of variants, we run this study with a reduced compute budget (shorter training and smaller splits) and report mean and standard deviation over three seeds; the emphasis is on consistent relative changes.

Table~\ref{tab:ablations} highlights four findings that sharpen the method narrative.
First, the robust log-distance kernels are essential: removing them and retaining only cosine similarities (cos\_only) collapses performance to near-random, while using only the log-distance family (log\_only) improves AUPRC by approximately $+0.031$ relative to the six-channel baseline, suggesting that magnitude-sensitive but robust pairwise structure is the dominant signal in this benchmark.

\begin{figure}[t]
\centering
\includegraphics[width=\linewidth]{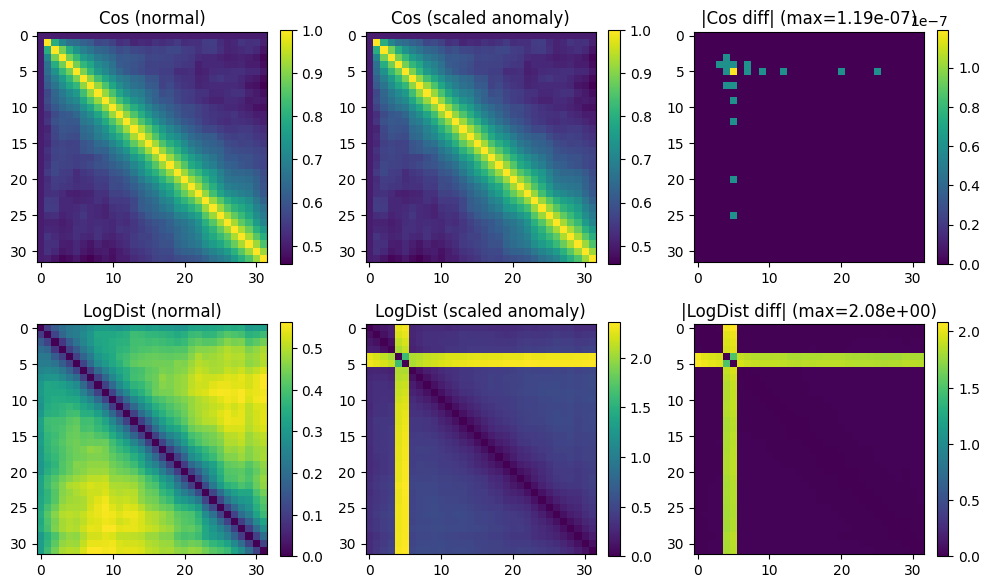}
\caption{Diagnostic illustrating why cosine-only kernel images can collapse under magnitude-driven anomalies. A short-segment amplitude scaling leaves the cosine kernel essentially unchanged (top row; max difference $\approx 10^{-7}$), while the robust log-distance kernel changes substantially (bottom row; max difference $\approx 2.08$), producing high-contrast structure. This supports the ablation result that cosine-only channels are ineffective in our setting, whereas log-distance channels remain discriminative. A controlled diagnostic in the appendix (Table~\ref{tab:cos-log-diagnostic}) confirms this mechanism: cosine-kernel statistics are near-random for amplitude scaling and additive magnitude perturbations, whereas log-distance statistics separate normal and anomalous windows almost perfectly.}
\label{fig:cos-collapse-diagnostic}
\end{figure}

\paragraph{Why cosine-only collapses.}
Cosine similarity is invariant to per-time rescaling: for any $\alpha>0$, $\cos(\alpha z_i,z_j)=\cos(z_i,z_j)$, so magnitude-driven anomalies can be suppressed when kernels are built after $\ell_2$ normalization.
In high-dimensional sketches, cosine similarities also tend to concentrate, yielding low-contrast kernel matrices that are easy to reconstruct and therefore provide weak anomaly signal.
Figure~\ref{fig:cos-collapse-diagnostic} visualizes this effect directly: a short amplitude scaling barely changes the cosine kernel (difference $\approx 10^{-7}$), whereas the robust log-distance kernel exhibits a large structured change.

Second, temporal differencing matters: removing the signed difference channels (no\_delta\_only) degrades AUPRC, whereas removing only the absolute-difference channels (no\_absdelta\_only) has a smaller effect, consistent with $\Delta g$ carrying discriminative signed dynamics.
Third, kernel-aware structured masking contributes: restricting masks to random symmetric patterns (mask\_random\_only) consistently reduces AUPRC, supporting the use of row/column, band, and block masks that force the model to reason across multiple kernel regions.
Fourth, representation normalization is important: removing the $\sqrt{n_t}$ normalization (no\_sqrt\_nt\_normalization) decreases performance and increases variance, indicating that controlling for time-varying observed counts is beneficial when $C$ and missingness vary.

Two additional ablations indicate that some components are less critical at $L{=}32$ in this synthetic setting: removing lag-based positional encoding (no\_lag\_positional) or the scale token (no\_scale\_token) changes performance only slightly.
Finally, removing the presence stream (no\_presence\_stream) slightly improves AUPRC in this benchmark, which is consistent with the fact that missingness is generated independently of anomalies; in settings where missingness is structured or correlated with abnormal events, the presence stream is expected to become more informative.

\begin{table}[t]
\centering
\scriptsize
\caption{Component ablation under holdout\_C at anomaly rate 0.10 (mean $\pm$ std over 3 seeds). The baseline corresponds to the six-channel hybrid image with saturating presence scaling, $\sqrt{n_t}$ normalization, lag-aware positional encoding, the scale token, and the full kernel-aware masking mixture.}
\label{tab:ablations}
\begin{tabular}{lccc}
\toprule
Variant & AUPRC & AUROC & $\Delta$AUPRC vs. base \\
\midrule
log\_only & 0.6158$\pm$0.0358 & 0.7775$\pm$0.0265 & +0.0310 \\
no\_presence\_stream & 0.6114$\pm$0.0367 & 0.7620$\pm$0.0386 & +0.0265 \\
no\_absdelta\_only & 0.5872$\pm$0.0502 & 0.7482$\pm$0.0358 & +0.0024 \\
base & 0.5848$\pm$0.0506 & 0.7608$\pm$0.0383 & +0.0000 \\
no\_scale\_token & 0.5814$\pm$0.0540 & 0.7639$\pm$0.0407 & $-$0.0034 \\
no\_lag\_positional & 0.5789$\pm$0.0520 & 0.7603$\pm$0.0463 & $-$0.0060 \\
lambda\_linear\_no\_saturation & 0.5785$\pm$0.0398 & 0.7582$\pm$0.0119 & $-$0.0063 \\
no\_deltas\_and\_absdeltas & 0.5704$\pm$0.0501 & 0.7278$\pm$0.0464 & $-$0.0145 \\
no\_delta\_only & 0.5649$\pm$0.0501 & 0.7563$\pm$0.0373 & $-$0.0199 \\
mask\_random\_only & 0.5611$\pm$0.0458 & 0.7572$\pm$0.0436 & $-$0.0237 \\
no\_sqrt\_nt\_normalization & 0.5593$\pm$0.0806 & 0.7567$\pm$0.0349 & $-$0.0256 \\
cos\_only & 0.1216$\pm$0.0317 & 0.4876$\pm$0.0186 & $-$0.4632 \\
\bottomrule
\end{tabular}
\end{table}

\section{Conclusion}
We studied window-level anomaly detection under variable-cardinality multivariate time series with missingness, where the number of observed variables can vary across windows and may be unseen during training.
SMKC decouples dynamic input structure from detection by mapping raw windows into a fixed-width hashed sequence and a kernel image that exposes global temporal structure.
Masked reconstruction and teacher--student masked prediction provide complementary signals that improve robustness, and lightweight validation-time fusion further enhances performance.
Across both in-distribution and holdout-$C$ protocols, SMKC achieves the best overall AUPRC and improves low-false-alarm operating behavior.

A key practical result is that random-projection kNN on the SMKC representation yields a strong training-free detector, supporting cold-start deployment when training is infeasible.
A targeted collision sweep shows that performance saturates once hash collisions are modest, supporting moderate hash widths such as $m=64$--$128$.
Finally, a component ablation study clarifies which elements matter most in the synthetic benchmark: robust log-distance kernel channels and signed temporal differencing are principal contributors, kernel-aware structured masking improves learning, and per-time normalization by $\sqrt{n_t}$ stabilizes performance under dynamic observed counts.
These findings strengthen the conclusion that representation design is a primary lever for variable-cardinality anomaly detection, with self-supervised learning and calibration acting as effective refinements when resources permit.


\appendix
\section{Additional Diagnostics}
\label{sec:appendix-diagnostics}
\begin{table}[t]
\centering
\caption{Cosine vs.\ log-distance diagnostic illustrating scale-invariance of cosine similarity. Metrics are computed from simple kernel-summary statistics on balanced normal/anomalous sets.}
\label{tab:cos-log-diagnostic}
\begin{tabular}{lcccc}
\toprule
Task & Cosine AUROC & Cosine AUPRC & LogDist AUROC & LogDist AUPRC \\
\midrule
Amplitude scaling (short segment) & 0.511 & 0.513 & 1.000 & 1.000 \\
Additive perturbation (short segment) & 0.328 & 0.387 & 1.000 & 1.000 \\
\bottomrule
\end{tabular}
\end{table}

\end{document}